\documentclass[conference]{IEEEtran}
\IEEEoverridecommandlockouts
\usepackage{cite}
\usepackage{amsmath,amssymb,amsfonts}
\usepackage{algorithmic}
\usepackage{graphicx}
\usepackage{textcomp}
\usepackage{xcolor}
\usepackage{url}
\usepackage{graphics} % for pdf, bitmapped graphics files
\usepackage{epsfig} % for postscript graphics files
\usepackage{times} % assumes new font selection scheme installed
\usepackage{amsmath} % assumes amsmath package installed
\usepackage{amssymb}  % assumes amsmath package installed
\usepackage{float}
\usepackage{booktabs}
\usepackage{changepage}
\usepackage{multirow}
\usepackage{makecell}
\usepackage{pbox}
\usepackage{tabularray}
\usepackage[hidelinks]{hyperref}
\usepackage{balance}
\usepackage{bm}

\makeatletter
\newcommand{\newlineauthors}{%
  \end{@IEEEauthorhalign}\hfill\mbox{}\par
  \mbox{}\hfill\begin{@IEEEauthorhalign}
}
\makeatother

\def\BibTeX{{\rm B\kern-.05em{\sc i\kern-.025em b}\kern-.08em
    T\kern-.1667em\lower.7ex\hbox{E}\kern-.125emX}}
\begin{document}

\title{Unsupervised UAV 3D Trajectories Estimation with Sparse Point Clouds\\

\thanks{This work is supported by National Research Foundation, Singapore, under its Medium-Sized Center for Advanced Robotics Technology Innovation.}
}

\author{
\IEEEauthorblockN{Hanfang Liang}
\IEEEauthorblockA{
\textit{Jianghan University,}\\
China. \\
hanfangliang@stu.jhun.edu.cn}
\and
\IEEEauthorblockN{Yizhuo Yang}
\IEEEauthorblockA{
\textit{Nanyang Technological University,}\\
Singapore. \\
yizhuo001@e.ntu.edu.sg}
\and
\IEEEauthorblockN{Jinming Hu}
\IEEEauthorblockA{
\textit{Jianghan University,}\\
China. \\
18571861306@stu.jhun.edu.cn}
\newlineauthors
\IEEEauthorblockN{Jianfei Yang}
\IEEEauthorblockA{
\textit{Nanyang Technological University,}\\
Singapore. \\
jianfei.yang@ntu.edu.sg}
\and
\IEEEauthorblockN{Fen Liu}
\IEEEauthorblockA{
\textit{Nanyang Technological University,}\\
Singapore. \\
fen.liu@ntu.edu.sg}
\and
\IEEEauthorblockN{Shenghai Yuan*}
\IEEEauthorblockA{
\textit{Nanyang Technological University,}\\
Singapore. *Corresponding-\\
 Author: shyuan@ntu.edu.sg}
}

\maketitle

\begin{abstract}
Compact UAV systems, while advancing delivery and surveillance, pose significant security challenges due to their small size, which hinders detection by traditional methods. This paper presents a cost-effective, unsupervised UAV detection method using spatial-temporal sequence processing to fuse multiple LiDAR scans for accurate UAV tracking in real-world scenarios. Our approach segments point clouds into foreground and background, analyzes spatial-temporal data, and employs a scoring mechanism to enhance detection accuracy. Tested on a public dataset, our solution placed 4th in the CVPR 2024 UG2+ Challenge, demonstrating its practical effectiveness. We plan to open-source all designs, code and sample data for the research community @ \url{github.com/lianghanfang/UnLiDAR-UAV-Est}.
\end{abstract}

\section{Introduction}
\label{sec:intro}
Drones have revolutionized various industries by allowing precise fertilization in agriculture and allowing detailed inspection of hard-to-reach structures \cite{cao2021distributed,lyu2023vision,cao2020online,lyu2021vision,esfahani2021learning}.
However, the potential for malicious drone use is a major concern. They can be used for unauthorized surveillance \cite{lyu2022structure}, drug trafficking \cite{cao2022direct}, smuggling \cite{lyu2023spins}, and even the deployment of grenades in war zones. This threat highlights the urgent need for advanced detection systems to detect hostile drones effectively.

\begin{figure}[t]
\centering
\vspace{-10pt}
 \includegraphics[width=0.49\textwidth]{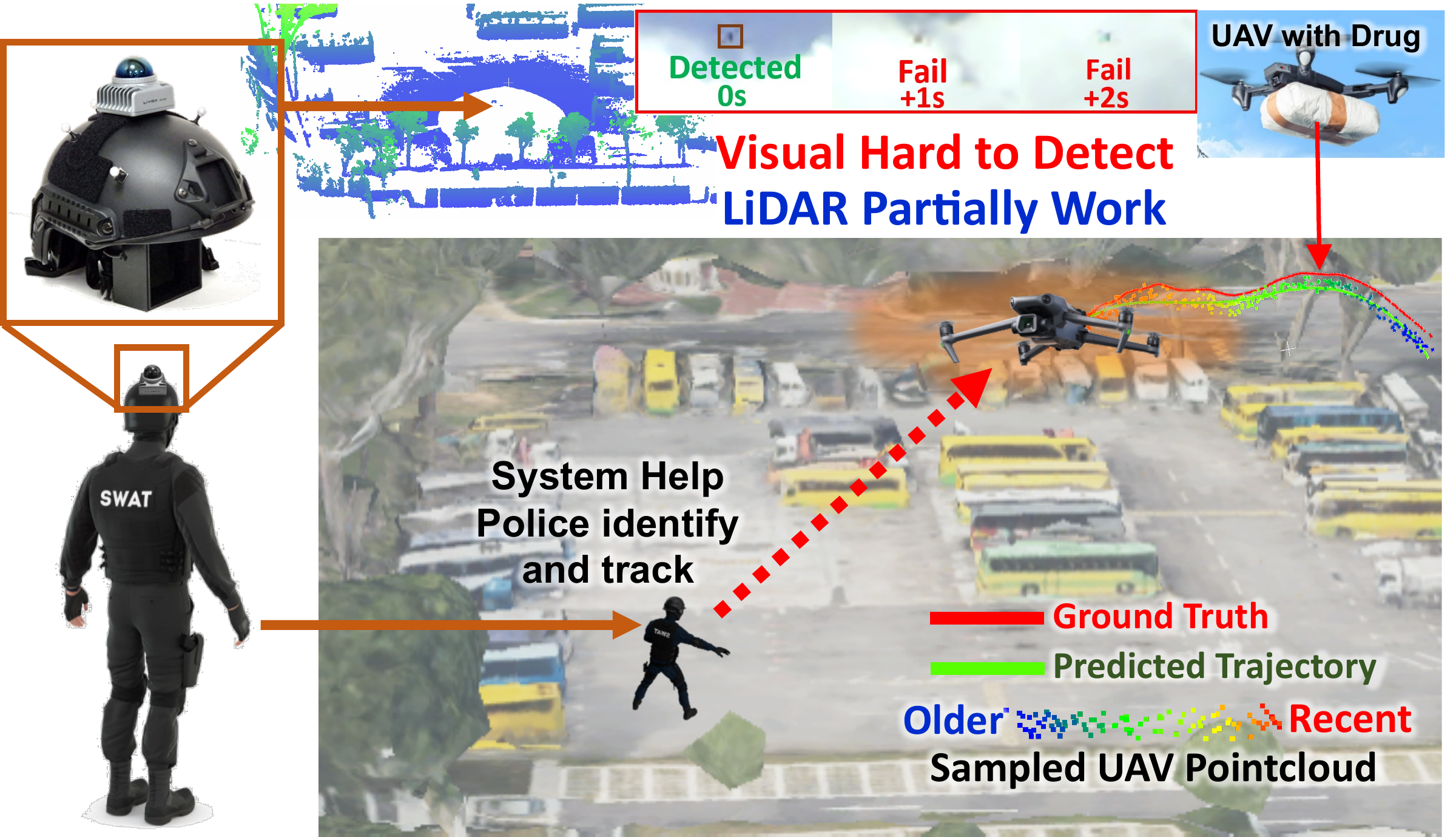}
 \vspace{-18pt}
 \tiny
\caption{Illustration of detecting and tracking compact drones using a single low-cost sparse LiDAR to identify threats.}
\vspace{-10pt}
\label{figure1illrustaraion}
\end{figure}

Detecting compact UAVs is challenging. Existing solutions rely on \textbf{UAV control signals} \cite{si2024application,wallace2019evaluating} to detect, but can be easily bypassed by changing frequencies, using 5G networks, or fully autonomous drones \cite{yuan2021survey,nguyen2021viral}. \textbf{Visual}-based methods \cite{coluccia2023drone,mistry2023drone,dong2023s} struggle with small objects at high altitudes. Narrow field-of-view cameras mounted on buildings can be operated manually to see the drone \cite{jiang2021anti,zhao21082nd,zhao20233rd,yuan2014Autonomous,wang2017heterogeneous,esfahani2018new,esfahani2019towards,esfahani2020local,esfahani2020unsupervised,yang2022overcoming,ji2022robust}, but this is impractical for field operations. Wide field of view cameras can monitor a larger area, but often only capture a few pixels of the drone as shown in Fig. \ref{figure1illrustaraion} . \textbf{Radar} can detect drones effectively, but cheaper models are noisy \cite{he2024detection} and expensive ones are expensive and power demanding \cite{wang2021deep}. \textbf{Audio}-based detection \cite{al2019audio,lei2024audioarraybased3duav,xiao2024tametemporalaudiobasedmamba,xiao2024avdtec} is intuitive but often less effective, with most commercial drones being very quiet at a distance. \textbf{LiDAR} can detect drones, but its data is sparse at long ranges \cite{vrba2023onboard}. In general, there is no perfect solution for drone detection.

This work aims to accurately detect drones regardless of their control signal frequency or autonomy, including small drones at high altitudes, without manual operation. It ensures practicality for wide field operations and affordability for use in a single person or a single vehicle, as shown in Fig. \ref{figure1illrustaraion}.

In this paper, we propose a concurrent clustering method for analyzing point clouds from a low-cost 3D LiDAR system. First, we perform global-local clustering to exclude large static objects. Then, we refine clustering using spatiotemporal density and voxel attributes to identify moving targets and isolate the UAV trajectory. Finally, we use spline fitting to reconstruct the UAV's spatial trajectory, enhancing detection accuracy, reducing noise, and eliminating irrelevant data for clearer insights into drone movements.

\begin{figure*}[ht]
\centering
\includegraphics[width=0.85\textwidth]{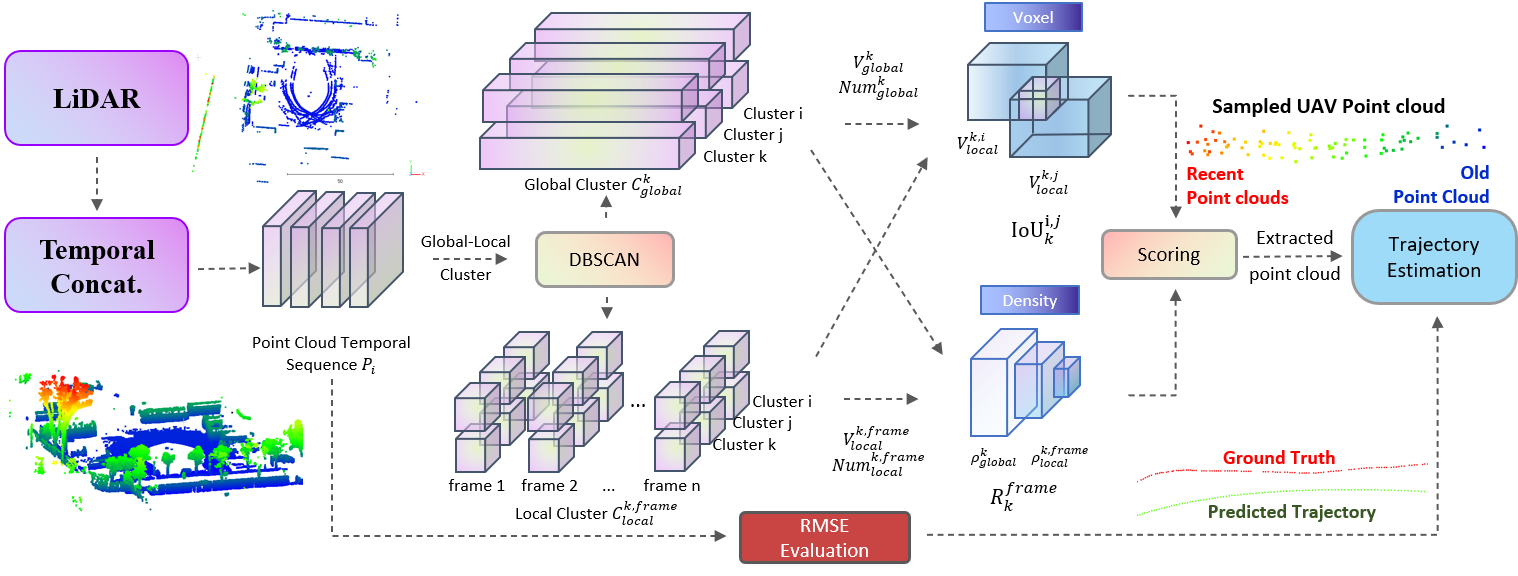}
\vspace{-15pt}
\caption{System Overview: Our algorithm uses DBSCAN to cluster point clouds, compares spatial-temporal changes, filters non-UAV data, and estimates UAV trajectories with spline fitting, measuring error with MSE.}
\label{fig:ICASSsystem}
\vspace{-15pt}
\end{figure*}

The main \textbf{contributions} of our work are as follows:
\begin{itemize}
\item \textbf{Unsupervised Trajectory Estimation:} We propose a fast, unsupervised method for detecting drone trajectories and positions from LiDAR point cloud data without any labels for supervised learning.
\item \textbf{Spatio-Temporal Analysis:} Our spatio-temporal voxel and density analysis method, with a scoring mechanism, isolates the correct trajectory point set.
\item \textbf{Extensive Benchmarking:} We benchmarked and tested various modalities with different methods to validate the performance of the system. To the best of our knowledge, this is the first benchmark of its kind for Anti-UAV study.
\item \textbf{Open-Source for All:} We plan to open-source our design, codes, scripts, and processed data for the benefit of the community and the general public \url{github.com/lianghanfang/UnLiDAR-UAV-Est}.
\end{itemize}
\textbf{International Recognition:} The proposed method \cite{liang2024separatingdronepointclouds} is an improved iteration of our award-winning solution from the CVPR 2024 UG2+ Challenge, enhancing its cost-effectiveness, robustness and reliability for practical field applications.

\vspace{-1.1em}
\section{Related Works}
\label{sec:Related}
\vspace{-0.5em}

This section reviews the limited literature on UAV detection and tracking, focusing on a few key approaches.

Vision-based UAV detection has evolved through deep learning, addressing challenges highlighted in studies such as Det-Fly \cite{zheng2021air} and MAV-VID, Drone-vs-Bird, and Anti-UAV \cite{isaac2021unmanned}. The methods have improved accuracy by augmenting the data and optimizing YOLOv4 for small UAV detection \cite{liu2021real}, and through transfer learning and adaptive fusion using simulated data \cite{rui2021comprehensive}.

Motion-assisted MAV detection integrates motion and appearance features using fixed and mobile cameras. Fixed camera methods employ background subtraction and CNN-based classification \cite{seidaliyeva2020real}, while mobile cameras utilize spatio-temporal characteristics \cite{xie2021small}-\cite{xie2020adaptive}, but can struggle in dynamic environments. Another approach combines appearance and motion-based classification to distinguish MAVs from distractors \cite{zheng2022detection}, albeit facing challenges with similar moving objects.

Detection from moving cameras is complex due to the background and target motion mixing together. Methods using UAV-to-UAV datasets and hybrid classifiers \cite{li2021fast} contend with background interference. Two-stage segmentation and feature super-resolution \cite{ashraf2021dogfight, wang2023low} offer advancements but grapple with issues like motion blur and occlusions in complex settings.

LiDAR systems, widely used for object detection and tracking, face unique challenges with UAVs due to their small size, shape variability, diverse materials, high speeds, and unpredictable movements. One method adjusts the integration time of the LiDAR frame based on drone speed and distance to improve the density and size of the point cloud, but this approach is intricate and sensitive to parameter settings \cite{qingqing2021adaptive}. Another strategy reduces LiDAR beams with probabilistic analysis and repositions the sensor for wider coverage, yet it struggles with continuous tracking of small points \cite{dogru2022drone}.
\begin{figure*}[ht]
\centering
\small 
\includegraphics[width=0.88\textwidth]{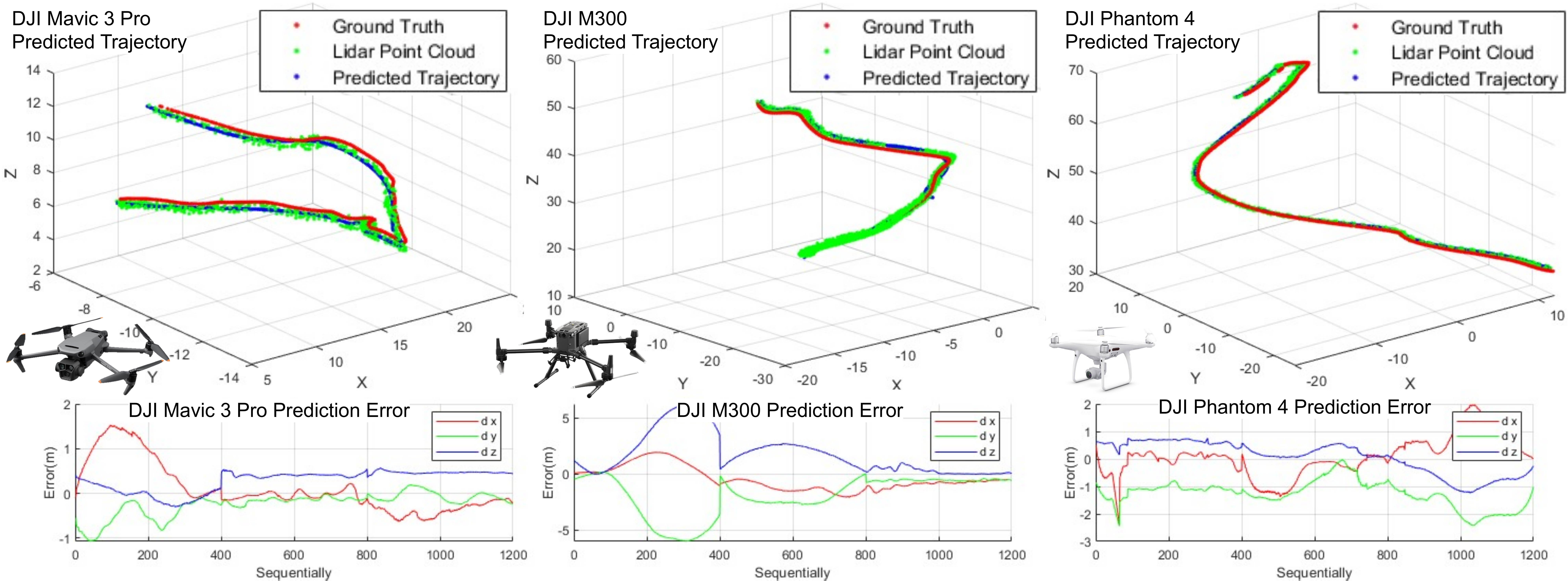}
\vspace{-11pt}
\small
\caption{This figure shows sampled points, ground truth, and our predicted trajectory, showing the accuracy of our solution.}
\label{fig:two_rows_three_images}
\vspace{-15pt}
\end{figure*}

Segmentation methods combined with object models and temporal information improve the effectiveness of UAV detection and tracking effectiveness, though they are constrained by segmentation and object model accuracy \cite{razlaw2019detection}. Euclidean distance clustering and particle filtering algorithms offer accurate yet computationally efficient solutions, albeit sensitive to data noise and outliers \cite{wang2021study}. %A novel approach treats LiDAR as a camera, integrating image and point cloud data for enhanced UAV detection using computer vision techniques, demanding substantial computational resources and advanced training data \cite{sier2023uav}.
In summary, while several methods address the challenges of UAV detection and tracking with LiDAR, each method presents distinct limitations and complexities, underscoring the need for ongoing research and development in this domain.
\vspace{-3pt}
\section{Proposed Methods}
\vspace{-3pt}
This section outlines our unsupervised spatial-temporal approach based on clustering to detect MAVs under challenging conditions. An overview of the system is shown in Fig. \ref{fig:ICASSsystem}.
\vspace{-03pt}
\subsection{Global-Local Point Set clusterings}
\vspace{-3pt}
Let $\mathcal{F}$ denote a sequence of LiDAR scan frames, with  $f$ denoting the number of frames.
$ \mathcal{P}$ represents a set of 3D points from a single scan in $\mathcal{F}$ . The number of points in set $\mathcal{P}$ is denoted $ card\, \mathcal{P}$. $\mathcal{C}$ denotes a cluster (subset) of points from $ \mathcal{P} $ and $\rho$ denotes the density of points, $\mathcal{V}$ denotes the voxels of a set of points.

For local representation, $ \left( \mathcal{P} | \mathcal{F}_{i}\right) $ represents the set of points $\mathcal{P}$ within the $i$-th frame $\mathcal{F}$. $\mathcal{C}_{j}\left( \mathcal{P} | \mathcal{F}_{i}\right)$ denotes the $j$-th cluster category of points from $ \left( \mathcal{P} | \mathcal{F}_{i}\right) $. For global representation, $ \sum_{i}^{j} \mathcal{F}$ denotes frame sequences from frame $i$ to frame $j$. $\mathcal{C}_{k}\left( \mathcal{P} | \sum_{i}^{j} \mathcal{F} \right)$ denotes the $k$-th category of the cluster after merging the points from $ \sum_{i}^{j} \mathcal{F}$. 
 
To distinguish between the results of global and local clustering, \( \mathcal{C}_{k}\left(\mathcal{P}, \mathcal{F}_{n} \mid\sum_{i}^{j}  \mathcal{F}\right) \) represents the \( k \)-th cluster of points in the \( n \)-th frame, where the clustering is derived from the sequence of frames \( \sum_{i}^{j}  \mathcal{F} \). 
 
 And $\mathcal{V}\left ( \mathcal{C} \right)$ indicates the size of the voxel occupied by cluster $\mathcal{C}$ in the space. Let $ \rho_{\mathcal{C}}$ be an operator that denotes the density of points in a cluster $\mathcal{C}$ derived from a set of points $\mathcal{P} $ in the context of a frame sequence $\mathcal{F} $.

We first superimpose the point cloud on the global time frames to obtain $ \left( \mathcal{P} |\sum_{i}^{j} \mathcal{F} \right) $, then use DBSCAN to perform clustering to obtain $\mathcal{C}\left(\mathcal{P} |\sum_{i}^{j} \mathcal{F}\right)$. Let $|\mathcal{C}|$ denote the cardinality of $\mathcal{C}$.  We calculate the density \( \rho_{\mathcal{C}}(\mathcal{P} \mid  \mathcal{F}) \)  of the point cloud in global point set.
\vspace{-0.5em}
\begin{equation}
\rho _{\mathcal{C}_{k}} \left( \mathcal{P} \mid \mathcal{F}\right)  = \frac{ | \, \mathcal{C}_{k}\left( \mathcal{P} \mid \mathcal{F} \right)| }{ \mathcal{V} \left( \mathcal{C}_{k}\left( \mathcal{P} \mid \mathcal{F}\right ) \right )}
\end{equation}

For the local point cluster, we first calculate the density \( \rho_{\mathcal{C}}(\mathcal{P} \mid \sum_{i}^{j} \mathcal{F}) \) of the point cloud in the point set $ \left( \mathcal{P} |\sum_{i}^{j} \mathcal{F} \right) $. And simultaneously calculate the spatial Intersection over Union (IoU) of the overlapping areas of voxels. Define the IoU of voxels in category $k$ of the cluster $\mathcal{C}_{k}\left(\mathcal{P} |\sum_{i}^{j} \mathcal{F}\right)$ between frame $i$ and $j$ as \(IoU_{k}^{i,j}\).
\vspace{-0.5em}
\begin{equation}
\rho_{\mathcal{C}_{k}}  \left( \mathcal{P} \mid \sum_{i}^{j} \mathcal{F}\right)  = 
  \frac{ | \,\mathcal{C}_{k}\left(\mathcal{P} |\sum_{i}^{j} \mathcal{F}\right) | }{ \mathcal{V} \left( \mathcal{C}_{k}\left( \mathcal{P} |\sum_{i}^{j} \mathcal{F}\right ) \right )}
\end{equation}
\vspace{-5pt}
\begin{equation}
IoU_{k}^{i,j} = \frac{\mathcal{V} \left( \mathcal{C}_{k}\left( \mathcal{P}|\mathcal{F}_{i}\right ) \right ) \cap \mathcal{V} \left( \mathcal{C}_{k}\left( \mathcal{P}|\mathcal{F}_{j}\right ) \right )}{\mathcal{V} \left( \mathcal{C}_{k}\left( \mathcal{P}|\mathcal{F}_{i}\right ) \right ) \cup \mathcal{V} \left( \mathcal{C}_{k}\left( \mathcal{P}|\mathcal{F}_{j}\right ) \right )} 
\end{equation}\vspace{-8pt}

And calculate the ratio of local density to global density as relative density \(\mathcal{R}_{k}^{i,j}\).
\vspace{-15pt}
\begin{equation}
\mathcal{R}_{k}^{i,j}=\frac{\rho\left ( \mathcal{C}_{k} \left( \mathcal{P}|\sum_{i}^{j} \mathcal{F}\right) \right)}{\rho\left ( \mathcal{C}_{k} \left( \mathcal{P}_{ \mathcal{F}}\right) \right)}
\end{equation}

At this point, through the global-local clustering, the relative density of each cluster point set \(\mathcal{R}_{k}^{i,j}\) and the IoU of voxels \(IoU_{k}^{i,j}\) can be obtained.

\begin{table*}
\footnotesize
\centering
\footnotesize
\caption{Benchmark for wide-area drone estimation of MMAUD V2 and V3 Challenging dataset}
%\vspace{-5pt}
\begin{tblr}{
  cells = {c},
  cell{1}{1} = {r=2}{},
  cell{1}{2} = {r=2}{},
  cell{1}{3} = {r=2}{},
  cell{1}{4} = {r=2}{},
  cell{1}{5} = {c=3}{},
  cell{1}{8} = {c=3}{},
  cell{1}{11} = {c=2}{},
  cell{1}{13} = {r=2}{},
  hline{1,3,6,8,15} = {-}{},
  hline{2} = {5-13}{},
}
\hline
\textbf{Methods }     & \textbf{Modality }    &    \textbf{Training}      &    \textbf{Bandwidth}        & \textbf{Day RMSE (m)} &             &             & \textbf{Night RMSE~ (m)} &             &             & \textbf{RMSE (m) } & &  $\overline{\textbf{ RMSE}  }$  \textbf{(m)}              \\
                      &         &       &    & \textbf{Dx}             & \textbf{Dy} & \textbf{Dz} & \textbf{Dx}               & \textbf{Dy} & \textbf{Dz} & \textbf{Day}       & \textbf{Night} \\
\textbf{VisualNet \cite{yang2024av}}    & Visual       &  Supervised   & 73.7Mpt/s & \underline{0.24}                    &\textbf{0.39}        & \underline{0.32}        & 1.98                      & 6.10        & 8.13        & \underline{0.65}               & 11.45           &      6.05     \\
\textbf{DarkNet \cite{liu2021real}}      & Visual       &   Supervised  & 73.7Mpt/s & \textbf{0.23}                    & \underline{0.46}        & \textbf{0.23}        & 1.84                      & 5.50        & 4.57        & \textbf{0.63}               & 8.31      &  4.47     \\
\textbf{YOLOv5s \cite{jocher2022ultralytics}}      & Visual       &   Supervised  & 73.7Mpt/s & 0.46                    & 0.57        & 1.04        & \textbf{0.64}                      & 1.76        & 1.59        & 1.27               & 4.71       & 2.99   \\
\textbf{AudioNet \cite{yang2024av}}     & Audio        &    Supervised & 0.18MHz & 0.60                    & 1.76        & 1.59        & 0.60                      & 1.76        & 1.59        & 2.80               & 2.80 & 2.80           \\
\textbf{VorasNet \cite{vora2023dronechase}}     & Audio       &   Supervised   & 0.18MHz & 0.54                    & 1.59        & 1.51        & 0.54                      & \underline{1.59}        &  \underline{1.51}        & 2.64               & \underline{2.64}       &\underline{2.64}    \\
\textbf{VoxelNet \cite{zhou2018voxelnet}}   & LiDAR       &   Supervised   & 0.20Mpt/s & 6.37                   & 7.75        & 5.89       & 6.37                       & 7.75        & 5.89        & 11.63              & 11.63     & 11.63     \\
\textbf{PointNet \cite{qi2017pointnet}}     & LiDAR       &   Supervised & 0.20Mpt/s   & /                    & /        & /        & /                      & /        & /        & 76.47               & 76.47       & 76.47    \\
\textbf{PointPillars \cite{lang2019pointpillars}} & LiDAR       &   Supervised & 0.20Mpt/s   & 4.34                    & 5.34        & 6.02        & 4.34                      & 5.34        & 6.02        & 9.14               & 9.14      & 9.14     \\
\textbf{VoteNet \cite{ding2019votenet}}      & LiDAR       &   Supervised  & 0.20Mpt/s  & /                    & /        & /        & /                      & /        & /        & 104.38               & 104.38    & 104.38        \\
\textbf{SECOND\cite{yan2018second}}    & LiDAR       &   Supervised & 0.20Mpt/s   & 5.05                    & 6.04        & 5.71       & 5.05                      & 6.04        & 5.71        & 9.73               & 9.73         & 9.73     \\
\textbf{SPVNAS \cite{tang2020searching}}       & LiDAR       &   Supervised & 0.20Mpt/s   & 2.24                   & 4.99       & 3.84        & 2.24                      & 4.99        & 3.84        & 6.69               & 6.69        & 6.69    \\
\textbf{Ours}         & LiDAR       &   Unsupervised  & 0.20Mpt/s  & 0.72                    & 0.85        & 0.76        & \underline{0.72}                      & \textbf{0.85}        & \textbf{0.76}        & 1.35               & \textbf{1.35}       & \textbf{1.35}                  
\end{tblr}
	\vspace{1pt}
	\begin{adjustwidth}{0cm}{0cm}
        
	\footnotesize{ RMSE: Error between predicted and actual values. The smaller, the better the estimation.  $\overline{\textbf{ RMSE} }$ denotes average error between day and night. }\\
	\footnotesize{ Mpt/s denotes Mega Sampling Points Per Second of input. ``/" denotes fails to detect. Best results are \textbf{boldened}, and second-best results are \underline{underlined}}
	\end{adjustwidth}
	\label{tab:example}
 \vspace{-15pt}
\end{table*}

\subsection{Scoring Mechanism and Trajectory Prediction}
For moving objects, the positions of the voxels change over time frames, causing a lower alignment compared to that of stationary objects. Stationary surfaces show an increase in point cloud density over time, while moving objects maintain a consistent density. We propose a scoring mechanism based on these density and voxel shifts, using a logarithmic function to stabilize and scale voxel IoU.

We define a voxel coincidence score for cluster k between local point set frame $\mathcal{C}\left( \mathcal{P}|\mathcal{F}_{i}\right)$ and $\mathcal{C}\left( \mathcal{P}| \mathcal{F}_{j}\right)$ as \(\bm{\psi} _{IoU}^{k}\).Define the score of point set density matching between $ \rho\left ( \mathcal{C}\left( \mathcal{P}| \mathcal{F}\right) \right)$ and $ \rho\left ( \mathcal{C}\left( \mathcal{P}|\sum_{i}^{j} \mathcal{F}\right) \right)$ as \(\bm{\psi}_{\rho}^{k}\).
\vspace{-0.5em}
\begin{equation}
\bm{\psi} _{IoU}^{k} = \sum_{k=1}^{n}log^{\frac{1}{IoU_{k}^{i,j}}}\
,
\quad \bm{\psi} _{\rho}^{k} = \sum_{k=1}^{n} e^{\mathcal{R}_{k}^{i,j}} 
\end{equation}
\begin{equation}
\bm{\psi}^{k} = \bm{\psi}_{\rho}^{k} + \lambda \times \bm{\psi}_{IoU}^{k}
\end{equation}

% The total score formula is recorded as \(Score^{k}\), where \(\lambda\) is  a hyperparameter set by us.

Based on the proposed scoring scheme, the category with the highest score \(\bm{\psi}^{k}\) can be identified as the final target with the highest confidence.

For the final trajectory based on the time frame, we use spline fitting on the UAV point cloud and interpolate based on the time frame to determine the spatial position at the corresponding time points.
 Define the cloud frame of the kth point after segmenting the background as \(P_{s}^{k}\). Sort the point clouds of each time frame according to the timestamp and merge them into a set of points \(\mathbb{P}_{uav}=\left \{ P_{s}^{0}, P_{s}^{1},...,P_{s}^{k}\right \}\). Among them, the points in the set of points \(\mathbb{P}_{uav}\) are selected as control points, and the three-dimensional spline S(u) can be expressed as:
\vspace{-0.8em}
\begin{equation}
S\left (  u\right )= \sum_{j=0}^{k}\sum_{i=0}^{n}P_{s}^{j}\left ( i \right )B_{i}\left ( u \right )
\end{equation}
Where \(B_{i}\) is the basis function of the spline. The three-dimensional curve is interpolated and fitted in the order of time frames to obtain the UAV spatial coordinates of the required time nodes.
\vspace{-1.1em}
\section{Experiment}
\vspace{-0.5em}
\subsection{Dataset}
\vspace{-0.3em}
We evaluated our algorithm on the difficult part of the MMAUD \cite{yuan2024MMAUD}, namely MMAUD\_v2 and MMAUD\_v3 sequences, featuring visual, LiDAR array, RADAR, and audio array sensors, with over 1700 seconds of multi-modal data in 50 sequences. Each sequence includes millions of sampling points of visual, LiDAR, audio, and radar data. For MMAUD\_V1 sequences, most detections are easy as UAVs typically fly within 40 meters. However, for MMAUD\_v2 and MMAUD\_v3 sequences, the 100-meter range makes smaller UAVs harder to detect with LIDAR.
\vspace{-0.5pt}
\subsection{Evaluation Metrics}
\vspace{-0.5pt}
We evaluate our algorithm using RMSE error, which directly evaluates system prediction accuracy in various conditions. By varying the lighting conditions, we can better understand the performance of each baseline method. The overall visual performance can be seen from Fig. \ref{fig:two_rows_three_images}, where green represents drone trajectories segmented through global and local clustering, red denotes actual drone positions, and blue indicates predicted positions. Our approach excels in noise reduction and precise drone trajectory extraction from point cloud data.
\vspace{-1.1em}
\subsection{Result and Discussion}
\vspace{-0.5pt}
The proposed solution demonstrates robust performance under various lighting conditions, as shown in table \ref{tab:example}. Traditional supervised LiDAR-based methods often expect dense data with large object sizes and end up with some of the worst performance due to sparse LiDAR data reflected by compact UAVs. Visual-based approaches perform well during the day with denser sampling points but exhibit significant performance drops at night. Audio-based methods show consistent performance day and night, but the overall accuracy is low. 

Our proposed solution manages to perform robust drone pose estimation for both day and night, even with very sparse point clouds. This shows that it is a practical solution for early warning applications of UAVs.
\vspace{-0.8em}
\section{Conclusion And Future Works}
\vspace{-0.5em}
This paper introduces an unsupervised approach for robust ground-based UAV detection using spatial-temporal and global-local clustering of sparse point cloud sequences. Our method extracts precise UAV trajectories from sparse and noisy data. We plan to open-source our design, codes, scripts, and sampled data. In future work, we aim to integrate active countermeasures, leveraging UAVs or EMP devices, to effectively neutralize drone threats using proposed perception inputs.

\vspace{-1.8em}
% Below is an example of how to insert images. Delete the ``\vspace'' line,
% uncomment the preceding line ``\centerline...'' and replace ``imageX.ps''
% with a suitable PostScript file name.
% -------------------------------------------------------------------------

\footnotesize
\bibliographystyle{IEEEbib}
\bibliography{mybib}

\end{document}